%% file: main.tex
\documentclass[letterpaper, 10pt, conference]{ieeeconf}

\IEEEoverridecommandlockouts  

\usepackage{comment}
\usepackage{amsmath}

\usepackage{graphicx}
\usepackage{rotating}
\usepackage{float}
\usepackage{mathrsfs}
\usepackage{amssymb}
\usepackage{autobreak}
\usepackage{mathtools}
\usepackage{wrapfig}
\usepackage{bbm}
\usepackage{algorithm}
\usepackage{graphicx, subcaption}
\usepackage{makecell}
\usepackage{booktabs}
\usepackage[svgnames, table]{xcolor}
\usepackage[normalem]{ulem} 
\usepackage{gensymb}
\usepackage{tabularray}
\UseTblrLibrary{booktabs}

\usepackage[left=0.75in, right=0.75in, top=0.75in, bottom=0.78in]{geometry}

\usepackage[]{algpseudocode}

\usepackage{lipsum}
\usepackage{outlines}
\usepackage{blindtext}
\usepackage{multicol}
\usepackage{tikz}
\usetikzlibrary{shapes,backgrounds}
\usepackage{array}
\usepackage{wrapfig}
\usepackage{thmtools} 
\usepackage{thm-restate}
\usepackage{epsfig} 
\usepackage{amsmath} 
\usepackage{algorithm}
\usepackage[]{algpseudocode}
\usepackage{tablefootnote}
\usepackage{multirow}
\usepackage{listings}
\usepackage{soul}

\usepackage{glossaries-extra}
\setabbreviationstyle[acronym]{long-short}
\glssetcategoryattribute{acronym}{nohyper}{true}

\newacronym{slam}{SLAM}{Simultaneous Localization and Mapping}
\newacronym{ba}{BA}{Bundle Adjustment}
\newacronym{sfm}{SfM}{Structure from Motion}
\newacronym{pgo}{PGO}{Pose-Graph Optimization}
\newacronym{vpr}{VPR}{Visual Place Recognition}
\newacronym{sgd}{SGD}{Stochastic Gradient Descent}
\newacronym{ils}{ILS}{Iterative Least-Squares}
\newacronym{gn}{GN}{Gauss-Newton}
\newacronym{lm}{LM}{Levenberg-Marquardt}
\newacronym{sdp}{SDP}{Semi-Definite Programming}
\newacronym{vo}{VO}{Visual Odometry}
\newacronym{vio}{VIO}{Visual-Inertial Odometry}
\newacronym{imu}{IMU}{Inertial Measurement Unit}
\newacronym{pnp}{PnP}{Perspective-n-Point}
\newacronym{dof}{DoF}{Degrees of Freedom}
\newacronym{ar}{AR}{Augmented Reality}
\newacronym{sota}{SOTA}{state-of-the-art}
\newacronym{rpr}{RPR}{Relative Pose Regression}
\newacronym{tsdf}{TSDF}{Truncated Signed Distance Field}
\newacronym{esdf}{ESDF}{Euclidean Signed Distance Field}
\newacronym{gvd}{GVD}{Generalized Voronoi Diagram}
\newacronym{gnc}{GNC}{Graduated Non-Convexity}
\newacronym{gcnn}{GCNN}{Graph Convolutional Neural Network}
\newacronym{llm}{LLM}{Large Language Model}
\newacronym{vlms}{VLMs}{Vision-Language Models}
\newacronym{eqa}{EQA}{Embodied Question Answering}
\newacronym{vqa}{VQA}{Visual Question Answering}
\newacronym{wfd}{WFD}{Wavefront Frontier Detection}

\usepackage{etoolbox}
\AtBeginEnvironment{algorithmic}{\everymath{\small}}

\def\sota{\gls{sota} }

\def\imu{\gls{imu}}

\def\tsdf{\gls{tsdf}}

\def\gvd{\gls{gvd} }

\def\llm{\gls{llm}}
\def\vlms{\gls{vlms}}
\def\eqa{\gls{eqa}}
\def\vqa{\gls{vqa}}
\def\wfd{\gls{wfd}}

\definecolor{revisionColor}{HTML}{FF0000}
\definecolor{backcolour}{rgb}{0.95,0.95,0.92}

\lstdefinestyle{ieeeprompt}{
  basicstyle=\ttfamily\small,
  breaklines=true,
  columns=fixed
  columns=fullflexible,
  backgroundcolor=\color{backcolour}
}

\input{macros.tex}
\title{Hierarchical Floorplan-Guided Vision-Language Exploration \\ for Embodied Question Answering}

\author{Albert Gassol Puigjaner, Kostas Alexis
\thanks{Autonomous Robots Lab, Norwegian University of Science and Technology (NTNU), Trondheim, Norway,
    {\tt \footnotesize
        \href{mailto:albert.g.puigjaner@ntnu.no}{albert.g.puigjaner@ntnu.no}}
    }
    \thanks{This work was supported in part by the Research Council of Norway under Grant NCEI (No. 357451) and in part by the European Commission under the Horizon Europe Programme through Grant SYNERGISE (No. 101121321).}
}

\definecolor{bestblue}{HTML}{D9EAF7}
\definecolor{secondorange}{HTML}{FCE4D6}
\definecolor{correctgreen}{HTML}{8FFAAA}
\newcommand{\best}[1]{\cellcolor{bestblue}\textbf{#1}}
\newcommand{\second}[1]{\cellcolor{secondorange}\textbf{#1}}

\newcommand{\vtext}[1]{%
  \begingroup
  \renewcommand{\arraystretch}{0.78}%
  \begin{tabular}[c]{@{}l@{}}#1\end{tabular}%
\endgroup
}
\definecolor{lightblue}{rgb}{0.12,0.49,0.85}
\makeatletter
\let\NAT@parse\undefined
\makeatother
\usepackage[colorlinks=true,linkcolor=blue,citecolor=blue,urlcolor=blue]{hyperref} 
\usepackage[capitalise]{cleveref}

\begin{document}

\maketitle

\begin{abstract}
\eqa~requires an agent to explore a previously unseen environment, gather relevant information, and answer questions about the scene. Recent approaches leverage \vlms~together with semantic maps or scene graphs to guide exploration. However, exploration is typically driven only by local observations, while structural priors about the environment remain largely unused. We propose \method, a hierarchical \eqa~framework that combines online scene graph construction, \vlm-based planning, semantic frontier exploration, and floorplan priors. The system incrementally builds a hierarchical scene graph and an open-vocabulary occupancy map from RGB-D observations, enabling a \vlm~to jointly reason over the scene graph, task-relevant visual observations, exploration history, and an estimated topological floorplan. Furthermore, we introduce a room-discovery strategy that leverages the floorplan and open-vocabulary frontier semantics to guide exploration toward semantically relevant yet currently unobserved room types. We evaluate \method~on the OpenEQA and ExploreEQA benchmarks and demonstrate deployment on a quadruped robot in real indoor environments. Our results demonstrate the benefit of combining \vlm-based hierarchical planning with structural floorplan priors for the \eqa~task.
\end{abstract}

\setlength{\textfloatsep}{5pt}
\input{1-intro}
\input{2-related_work}

\input{3-problem_statement}
\input{4-method}
\input{5-experimental_results}

\input{6-conclusions}

\bibliographystyle{IEEEtran}
\bibliography{IEEEabrv, bibligraphy}

\end{document}

%% file: macros.tex
\input{notation}

\newcommand{\mypar}[1]{\noindent\textbf{#1}.}
\newcommand{\method}{\textsc{\small{HFLEX-EQA}}}
\newcommand{\vlm}{VLM}

\newcommand{\level}{L}
\newcommand{\mesh}{\level_m}
\newcommand{\object}{\level_o}
\newcommand{\navgraph}{\level_n}
\newcommand{\frontier}{\level_f}
\newcommand{\room}{\level_r}

\newcommand{\graph}{\cG}
\newcommand{\floorplan}{\cF}
\newcommand{\nodes}[2]{\cN_{#1}^{#2}}
\newcommand{\edges}[2]{\cE_{#1}^{#2}}

\newcommand{\visscore}[2]{\rho_{#1}(#2)}
\newcommand{\visweight}[1]{\lambda_{#1}}

\newcommand{\frontiergain}[2]{S_{#1}(#2)}

\newcommand{\class}{c}
\newcommand{\targetclass}{c^{\star}}
\newcommand{\weightclass}[1]{w(#1)}
\newcommand{\distscore}[3]{\operatorname{dist}_{#1}(#2,#3)}
\newcommand{\shortestdist}[2]{d(#1,#2)}
\newcommand{\classlabel}[1]{\ell(#1)}

%% file: notation.tex
\usepackage{amsopn}

\newcommand{\bI}{\mathbf{I}}

\newcommand\norm[1]{\left\lVert#1\right\rVert}

\newcommand{\cA}{\mathcal{A}}

\newcommand{\cC}{\mathcal{C}}
\newcommand{\cH}{\mathcal{H}}
\newcommand{\cN}{\mathcal{N}}

\newcommand{\cO}{\mathcal{O}}
\newcommand{\cF}{\mathcal{F}}

\newcommand{\cX}{\mathcal{X}}
\newcommand{\cI}{\mathcal{I}}

\newcommand{\cT}{\mathcal{T}}

\newcommand{\cG}{\mathcal{G}}

\newcommand{\cE}{\mathcal{E}}
\newcommand{\cU}{\mathcal{U}}
\newcommand{\cK}{\mathcal{K}}
\newcommand{\cV}{\mathcal{V}}
\newcommand{\cZ}{\mathcal{Z}}

\newcommand{\bb}{\mathbf{b}}

\newcommand{\bp}{\mathbf{p}}

\newcommand{\rmO}{\mathrm{O}}

\def\g2o{$g^2o$}
\def\t2v{\mathrm{t2v}}
\def\v2t{\mathrm{v2t}}
\def\ev2t{\mathrm{ev2t}}


%% file: 1-intro.tex

\glsreset{slam}
\glsreset{llm}
\glsreset{vlms}
\glsreset{eqa}
\section{INTRODUCTION}\label{sec:intro}

In \eqa~\cite{Das2018EQA}, an agent must actively explore an unknown environment to gather the necessary information to answer a specific question. Unlike conventional \vqa~\cite{Agrawal2015VQA}, the required visual evidence is not immediately available. Instead, the agent must decide where to explore, what observations to collect, and when sufficient evidence has been gathered to answer the question. Therefore, \eqa~requires semantic reasoning, memory mechanisms, and active exploration.

\vlms~have significantly improved the semantic reasoning capabilities of robotic agents via the knowledge acquired from internet-scale visual and textual data. Consequently, recent \eqa~systems~\cite{allen2024exploreeqa,saxena2025GraphEQA,OpenEQA2023} use \vlm s to reason over images or structured scene representations. However, in such methods, exploration is typically treated as a single planning problem, despite the different types of information that may be missing. For example, answering a question may require discovering a new room, obtaining a better view of an observed object, or further exploring a known part of the environment.

When the information needed to answer the question is in an unseen part of the environment, structural priors can be useful. Humans naturally use such priors in unfamiliar environments: when looking for a bed, one typically searches for a bedroom rather than exploring every room. Similar information is often available to robots through floorplans or building models, but existing \eqa~approaches do not exploit it to reason about which unseen rooms are question-relevant.

\begin{figure}[t]
\centering
\includegraphics[width=1.0\linewidth]{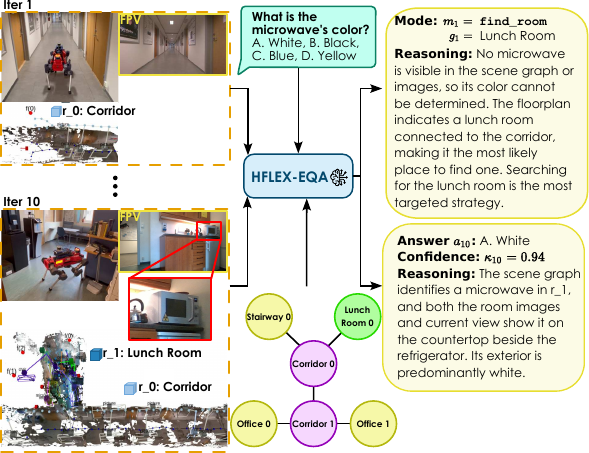}
\caption{Given the question \textit{``What is the microwave's color?''}, the robot must explore the unknown environment to find the required visual information. Using online RGB-D observations, the robot incrementally builds a scene graph and reasons over the visual memory and a topological floorplan to select the appropriate exploration strategy. After locating the lunch room and inspecting the microwave, it answers the question with high confidence.}
\label{fig:intro_example}
\vspace{-.7em}
\end{figure}

In this work, we propose Hierarchical Floorplan-guided Vision-Language Exploration for Embodied Question Answering (\method), a framework for active \eqa~that combines online scene graph construction with hierarchical \vlm-based planning. The system incrementally builds a hierarchical scene graph~\cite{hughes2022hydra}, while a high-level \vlm~planner reasons jointly over the scene graph, task-relevant visual observations, exploration history, and a floorplan prior to determine both the missing information and the appropriate exploration strategy. Dedicated low-level planners then perform room exploration, object inspection, or floorplan-guided room discovery. An example of our method's high-level reasoning and scene graph construction is shown in~\cref{fig:intro_example}. We evaluate our approach on the OpenEQA~\cite{OpenEQA2023} and ExploreEQA~\cite{allen2024exploreeqa} benchmarks and demonstrate deployment on a quadruped robot equipped with a custom sensing and compute payload.

The main contributions of this work are:

\begin{itemize}
\item We propose a hierarchical \vlm-based \eqa~framework consisting of a high-level reasoning module and specialized low-level planners that jointly reason over a 3D scene graph, relevant observations, and floorplan priors.

\item We introduce a floorplan-guided room-discovery strategy that combines topological floorplans with open-vocabulary frontier semantics to guide exploration toward semantically relevant unobserved room types. 

\item We evaluate the proposed approach on the OpenEQA and ExploreEQA benchmarks, where our method with floorplan priors outperforms existing \eqa~baselines. We additionally demonstrate real-world deployment on a quadruped robot in previously unseen environments.
\end{itemize}

In the remainder of this paper, we first review related work (\cref{sec:related}), followed by the problem formulation (\cref{sec:problem_statement}). We then describe the proposed method (\cref{sec:method}) and evaluate it in simulation and with real-world experiments (\cref{sec:results}). Conclusions are drawn in~\cref{sec:conclusions}.

%% file: 2-related_work.tex

\section{RELATED WORK}\label{sec:related}

\eqa~requires three key capabilities: maintaining a structured representation of the environment, efficiently exploring to gather relevant information, and reasoning over the observations. We therefore review related work on scene graphs, semantic exploration, and embodied question answering.

\mypar{3D Scene Graphs}
A key challenge in \eqa~is maintaining a relevant, structured, and lightweight scene representation to support semantic reasoning. 3D scene graphs have emerged as a compact representation for describing semantic abstractions and their relationships within 3D environments~\cite{Rosinol2021Kimera}. Compared to purely geometric or metric-semantic maps, scene graphs provide higher-level abstractions, such as rooms, objects and buildings, that facilitate semantic reasoning and scene understanding. Building upon this idea, Hydra~\cite{hughes2022hydra} introduced the first real-time framework for the incremental construction of hierarchical scene graphs. More recently, advances in \vlm s have enabled open-vocabulary scene graphs, both offline~\cite{Gu2024conceptgraphs,werby2023hovsg} and online~\cite{puigjaner2026reasoninggraph,Maggio2024Clio}, supporting language-grounded object search.

\mypar{Semantic Exploration}
Among the broad literature on semantic exploration, we focus on approaches that leverage \vlm s for open-vocabulary exploration. These methods address the problem of deciding where an agent should explore to efficiently gather task-relevant information. Early zero-shot approaches such as CLIP-Nav~\cite{dorbala2022clipnav} and Visual Language Maps~\cite{huang23vlmaps} leveraged \vlm s to score RGB images or construct open-vocabulary 2D grid maps to navigate toward language goals. Subsequent methods introduced open-vocabulary frontiers. In particular, VLFM~\cite{yokoyama2024vlfm} proposed scoring frontiers using the cosine similarity between the frontiers' vision-language features and language goals, while ESC~\cite{zhou2023esc} introduced soft commonsense constraints that model likely goal-to-object and goal-to-room proximity probabilities. Other approaches used a \llm~to score semantic frontiers~\cite{Yuan2024ZeroShot} or leveraged object-level scene graph representations~\cite{yin2024sgnav,Huang2026MSGNav}. On the other hand, CLIP on Wheels (CoWs)~\cite{gadre2022cow} established a benchmark for language-grounded zero-shot object navigation and proposed to use CLIP~\cite{Radford2021CLIP} to score frontiers. Our work builds upon these semantic exploration strategies by integrating open-vocabulary frontiers with hierarchical scene graph reasoning and floorplan priors.

\mypar{Embodied Question Answering}
Early \eqa~approaches~\cite{Das2018EQA} primarily relied on end-to-end learned navigation and \vqa~policies. More recently, OpenEQA~\cite{OpenEQA2023} introduced a benchmark for open-ended question answering in realistic 3D environments, showing the capabilities and limitations of modern \vlm s when provided with a complete sequence of images in the environment. ExploreEQA~\cite{allen2024exploreeqa} extended this setting to active exploration, proposing a \vlm-based agent that explores the environment and answers questions. GraphEQA~\cite{saxena2025GraphEQA} further demonstrated the benefits of grounding \vlm~planning with an online hierarchical scene graph augmented with semantic room labels, frontiers, and task-relevant visual memory. These works show that structured semantic memory and \vlm-based reasoning are key components for solving \eqa~tasks. However, exploration is typically guided by a single planning strategy using only locally observed information. In contrast, our approach introduces a hierarchical planning framework with specialized exploration modes and incorporates a topological floorplan to reason about semantically relevant but unobserved room types.

%% file: 3-problem_statement.tex
\section{PROBLEM STATEMENT}\label{sec:problem_statement}

We consider \eqa~in previously unseen indoor environments. The input consists of a question $q$, an optional set of answer choices $\cA$, and a stream of online RGB-D observations collected during exploration. The agent is also provided with an estimated floorplan graph,
\begin{equation}
\graph^{\floorplan}=(\nodes{}{\floorplan},\edges{}{\floorplan}),
\end{equation}
where nodes $\nodes{}{\floorplan}$ represent room instances annotated with semantic labels, and edges $\edges{}{\floorplan}$ encode room connectivity.

The objective is to actively explore the environment, gather the visual information required to answer $q$, and return:
\begin{equation}
a^*=F_{\text{EQA}}\left(q,\cZ_{0:t},\graph^{\floorplan}, \cA\right),
\end{equation}
where $\cZ_{0:t}$ are the RGB-D observations collected until time $t$ during exploration, $a^*$ is either one of the answer choices in $\cA$ or a free-form answer when no choices are provided and $F_{\text{EQA}}$ is our method. The floorplan graph $\graph^{\floorplan}$ serves as a structural prior to guide exploration. Since it is estimated rather than ground truth, it may be incomplete or inaccurate, containing missing or incorrectly classified room nodes as well as missing or incorrect connectivity edges.

%% file: 4-method.tex
\section{METHOD}\label{sec:method}

\begin{figure*}[t]
\centering
\includegraphics[width=\linewidth]{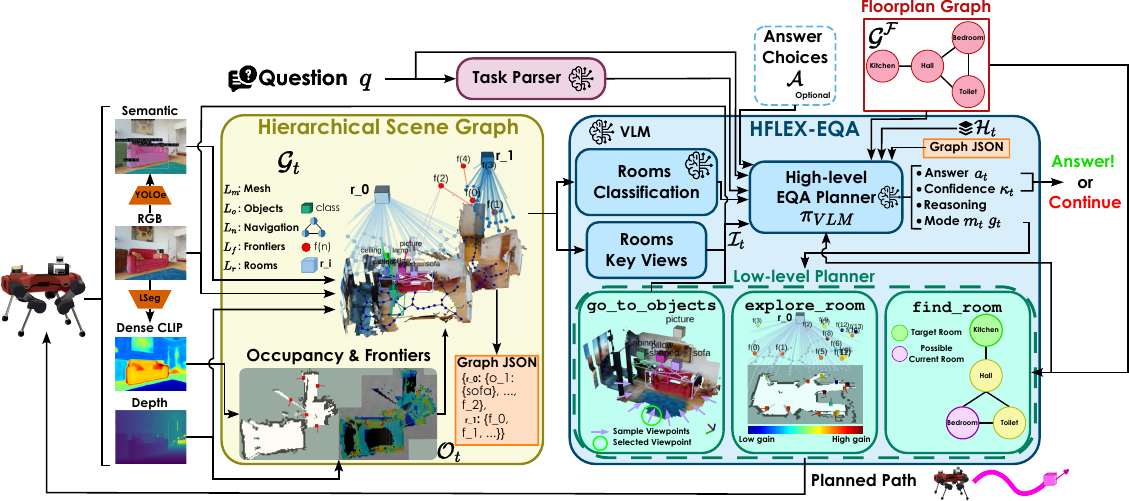}
\caption{\method~overview. Online RGB-D observations are used to incrementally construct a hierarchical scene graph $\graph_t$~\cite{hughes2022hydra} containing rooms $\room$, semantically-enhanced frontiers $\frontier$, a navigational graph $\navgraph$, objects $\object$ and a metric-semantic mesh $\mesh$. The question, optional answer choices, scene graph, relevant visual memory, action history, and floorplan are fed to a \vlm-based high-level planner $\pi_{VLM}$, which determines whether the question can be answered or further exploration is required. For exploration, the planner selects one of three modes $m_t$: (i) \textbf{\texttt{go\_to\_objects}}, (ii) \textbf{\texttt{explore\_room}}, or (iii) \textbf{\texttt{find\_room}}. A low-level semantic planner then selects frontier or viewpoint targets according to the mode and generates a path.}
\label{fig:overview}
\vspace{-5ex}
\end{figure*}

To address the active \eqa~task, we propose the hierarchical planning framework shown in~\cref{fig:overview} that combines semantic scene understanding, visual-language reasoning, and question-guided exploration. The agent incrementally constructs a semantic scene graph from online observations while leveraging an estimated floorplan prior to reason about the global structure of the environment. The high-level \vlm-based task planner selects the next exploration strategy and may answer once sufficient information has been collected, while specialized low-level planners convert its decisions into navigation goals.  We first introduce the scene graph representation and open-vocabulary occupancy mapping in~\cref{subsec:graph}, followed by the floorplan prior graph in~\cref{subsec:floorplan}, the high-level planning in~\cref{subsec:high-level} and the low-level planning in~\cref{subsec:low-level}. 

\subsection{Hierarchical Scene Graph}\label{subsec:graph}

At time $t$, we maintain an online hierarchical scene graph $\graph_t$. We use Hydra~\cite{hughes2022hydra} for scene reconstruction, object and room layer construction, and graph optimization. We augment it with semantic room labels, room key views, frontier nodes, and a navigation layer. The resulting graph nodes are:
\begin{equation}
\nodes{t}{} = \nodes{t}{\room} \cup \nodes{t}{\frontier} \cup \nodes{t}{\navgraph} \cup \nodes{t}{\object} \cup \nodes{t}{\mesh},
\end{equation}
corresponding to room ($\room$), frontier ($\frontier$), navigation ($\navgraph$), object ($\object$), and metric-semantic mesh ($\mesh$) layers. Each room node $r_i = (\class_{r_i}, \cK_i) \in \nodes{t}{\room}$ stores a semantic label $\class_{r_i}$ and a set of key views $\cK_i$. A key view $K \in \cK_i$ is added when its pose differs sufficiently in position and orientation and its CLIP~\cite{Radford2021CLIP} embedding is sufficiently different in cosine similarity to the previous key view. Room nodes are connected to the navigation nodes contained within them. Each frontier $f_i = (\bp_{f_i},\theta_{f_i},\phi_{f_i},s_{f_i})\in\nodes{t}{\frontier}$ represents a boundary between explored and unexplored space and stores its centroid $\bp_{f_i}$,  orientation toward unexplored space $\theta_{f_i}$, open-vocabulary semantic feature $\phi_{f_i}$, and size $s_{f_i}$. Frontiers are connected to the nearest navigation node and nearby objects, while navigation nodes encode traversability. Each object $o_i = (\bp_{o_i}, \bb_{o_i}, c_{o_i}, \cV_i) \in\nodes{t}{\object}$ stores its centroid $\bp_{o_i}$, bounding box $\bb_{o_i}$, semantic label $c_{o_i}$, and associated mesh vertices $\cV_i$. The mesh provides the dense representation from which the higher-level graph layers are constructed.

The graph is incrementally constructed from posed RGB-D observations using YOLOe~\cite{wang2025yoloe} for object detection. In parallel, we maintain an open-vocabulary 2D occupancy map $\cO_t$ using LSeg~\cite{li2022lseg} to extract pixel-wise CLIP features. These features are projected into a \tsdf~using depth and incrementally fused, with each cell storing its occupancy state and a running average of its semantic features. A \gvd~extracted from $\cO_t$ forms the navigation layer $\nodes{t}{\navgraph}$, while \wfd~detects connected boundaries between observed free and unexplored space to construct $\nodes{t}{\frontier}$. We compute the semantic feature $\phi_{f_i}$ of each frontier $f_i \in \nodes{t}{\frontier}$ by averaging the semantic features of occupancy-map cells within a fixed radius $r_f$ of its centroid $\bp_{f_i}$. 

\subsection{Floorplan Prior}\label{subsec:floorplan}

As described in~\cref{sec:problem_statement}, our method additionally leverages an estimated floorplan graph $\graph^{\floorplan}$. Unlike the scene graph $\graph_t$, which describes explored regions, the static floorplan prior provides a global topological graph representation of the environment: each room instance becomes a node with a semantic label, and edges connect adjacent rooms, as shown in~\cref{fig:overview}. Metric geometry and agent localization are not retained.

The floorplan is treated as a weak high-level structural prior rather than an accurate map. Thus, it is not assumed to be complete or fully accurate, and may contain missing rooms, incorrect room labels, or erroneous connectivity edges. Instead of requiring localization within the floorplan, our method uses the semantic room labels observed online together with the floorplan connectivity to reason about which room types are most likely to lead toward a question-relevant place. This avoids requiring exact floorplan localization while still exploiting its global structural information. Even an incomplete floorplan provides valuable global context that complements local semantic exploration. In our experiments, we construct $\graph^{\floorplan}$ from HM3D~\cite{Yadav2022HabitatMatterport3S} semantic annotations in simulation and manually from known room layouts in the real world.

\subsection{High-Level VLM Planner}\label{subsec:high-level}

Our high-level planner receives a compact representation:
\begin{equation}
\cX_t =
\left(q, \cA, \graph_t, \cI_t,
\cH_t, \graph^{\floorplan}\right),
\end{equation}
where $\cA$ is the optional set of answer choices, $\cI_t$ denotes the relevant visual memory, $\cH_t=\{(m_k,g_k)\}_{k=0}^{t-1}$ is the history of previous high-level decisions, and $\graph^{\floorplan}$ is the floorplan. At each planning iteration, our \vlm-based high-level planner maps this state to
\begin{equation}
\pi_{\vlm}(\cX_t) = \left(m_t, g_t, a_t, \kappa_t\right).
\label{eq:high_level_planner}
\end{equation}
$m_t \in \{\textbf{\texttt{explore\_room}}, \textbf{\texttt{go\_to\_objects}}, \textbf{\texttt{find\_room}}\}$ is the selected exploration mode and $g_t$ its corresponding target. Depending on $m_t$, $g_t$ is a room node $r$, one or more object nodes $o$, or a semantic room class $\class$. Finally, $a_t$ and $\kappa_t\in[0,1]$ are an optional answer and its confidence score. These outputs define the interface with the specialized low-level planners described in~\cref{subsec:low-level}.

The reasoning process of $\pi_{\vlm}$ consists of three stages. First, a relevant visual memory module selects task-relevant observations from the current view, room key views, and, when available, object inspection views, avoiding using the complete image observation history. Second, semantic room labels $\class_{r_i}$ are predicted from the room key views and objects. Finally, $\cX_t$ is provided to the \vlm-based task planner to select the next exploration mode and, when sufficient evidence is available, answer the question.

\mypar{Relevant visual memory}
Each room node $r_i\in\nodes{t}{\room}$ stores a set of key views $\cK_i$ and their corresponding CLIP image embeddings. For each room, at most $M$ key views are selected from $\cK_i$ according to their relevance to $q$, measured as the cosine similarity between the CLIP text embedding of the question and the CLIP image embedding of each key view. We denote the resulting set of selected images by $\cI_{r_i}$.

If the previous high-level decision is $m_{t-1}=\textbf{\texttt{go\_to\_objects}}$, the object inspection views $\cI_{\rmO}$ are included. Finally, the current observation $\bI_t$ is added, giving
\begin{equation}
\cI_t =
\left(\bigcup_{r_i\in\nodes{t}{\room}}
\cI_{r_i}\right)
\cup \cI_{\rmO} \cup \{\bI_t\},
\label{eq:visual_memory}
\end{equation}
where $\cI_{\rmO}$ is included only when object inspection views are available. This compact visual memory enables the planner to reason over the most informative images.

\mypar{Room label prediction}
Before querying the high-level planner, the semantic label $\class_{r_i}$ of each room node $r_i$ is predicted by a \vlm~using its key views $\cK_i$ together with the semantic labels of the object nodes contained in the room. We constrain the \vlm~to only predict classes present in the floorplan to allow the planner to relate observed rooms to the semantic room classes in the floorplan.

\mypar{\vlm-based high-level planner}
The compact state $\cX_t$ is provided to the task-planner \vlm~$\pi_{VLM}$, with $\graph_t$ and $\graph^{\floorplan}$ serialized as JSON. The prompt contains the graph representations, visual memory $\cI_t$, question $q$, history $\cH_t$, and, when available, answer choices $\cA$. The planner returns $(m_t,g_t,a_t,\kappa_t)$ according to~\cref{eq:high_level_planner}. If an answer $a_t$ is provided with confidence $\kappa_t\geq\kappa_{\mathrm{min}}$, it is returned as the final answer and the episode terminates. Otherwise, the low-level planner executes $(m_t,g_t)$ and exploration continues. We additionally prompt the \vlm~to describe the scene and floorplan graphs and provide an argument supporting its decision to strengthen its reasoning.

\subsection{Low-Level Planning}\label{subsec:low-level}

The role of the low-level planner is to execute the exploration strategy proposed by the high-level planner. Depending on the selected $m_t$, the agent may explore a known room, revisit observed objects, or search for a room type that has not yet been discovered. These modes operate on the room, frontier, and object nodes of $\graph_t$, using the attributes introduced in~\cref{subsec:graph} to select navigation goals.

\mypar{\texttt{explore\_room}}
This mode is executed when the high-level planner decides to explore a known target room $r_{target} \in \nodes{t}{\room}$, indicating that the information required to answer the question is likely located within that known but only partially explored target room. Exploration is therefore restricted to frontier nodes $f_i\in\nodes{t}{\frontier}$ associated with $r_{\mathrm{target}}$. To guide exploration within the room, we first query a task parser \vlm~using the \eqa~question and instruct it to predict a set of $N$ objects that are most likely to be relevant for answering it. These predicted object labels are converted into CLIP text embeddings, $\Psi(q)=\{\psi_j\}_{j=1}^{N}$. The semantic relevance of a frontier $f_i$ is computed as the maximum cosine similarity between its feature and the predicted object embeddings,
\begin{equation}
\frontiergain{\mathrm{exp}}{f_i}= \max_{j} \frac{\phi_{f_i}^\top\psi_j} {\|\phi_{f_i}\|\|\psi_j\|}.
\end{equation}

Intuitively, frontiers whose semantic appearance is more strongly associated with the predicted task-relevant objects receive higher scores. The agent navigates to the highest-scoring reachable frontier and orients itself according to its stored orientation $\theta_{f}$ toward the unexplored region

\mypar{\texttt{go\_to\_objects}} 
When the high-level planner identifies a previously observed object of the scene graph as relevant to the question, the objective is not simply to revisit the object but to obtain a more informative view of it. This is particularly important when answering the question requires observing object attributes, state or details that may not be visible from earlier observations.

To get this evidence, the planner searches for a viewpoint maximizing the expected visibility of the target object $o_{target}$. From its bounding box $\bb_{o_{target}}$, it estimates a minimum radius $r_{min}~=~\max(\delta_{o_{target}}^{xy}/\tan(\theta_x/2),\delta_{o_{target}}^z/\tan(\theta_y/2))$, where $\delta_{o_{target}}^{xy}$, $\delta_{o_{target}}^z$ are the horizontal and vertical bounding box ($\bb_{o_{target}}$) half-extents, and $\theta_x$, $\theta_y$ are the camera fields of view. The planner then samples azimuths and radii in $[r_{min},r_{min}(1+\epsilon)]$ around the centroid $\bp_{o_{target}}$, discarding viewpoints outside observed free space. Each candidate viewpoint $v$ is then evaluated according to
\begin{equation}
\frontiergain{\mathrm{vp}}{v} = \visweight{\mathrm{vis}}\visscore{\mathrm{vis}}{v} + \visweight{\mathrm{img}}\visscore{\mathrm{img}}{v}, \,\,\, \visweight{\mathrm{vis}} + \visweight{\mathrm{img}} = 1, 
\label{eq:visibility_score}
\end{equation}
where $\visscore{\mathrm{vis}}{v}$ measures the fraction of vertices in $\cV_{o_{target}}$ that remain visible after depth-based occlusion checking, and $\visscore{\mathrm{img}}{v}$ measures the fraction of object points projected inside the camera image. Intuitively, the scoring function favors viewpoints that maximize object visibility and coverage, increasing the likelihood that the attributes required to answer the question are observed. Other factors affecting view quality, such as viewing angle or illumination, are not explicitly modeled. The highest-scoring viewpoint is selected as the navigation goal. Once reached, the object view $\cI_{\rmO}$ is added to the relevant visual memory $\cI_t$.

\begin{figure}[t]
\centering
\includegraphics[width=\columnwidth]{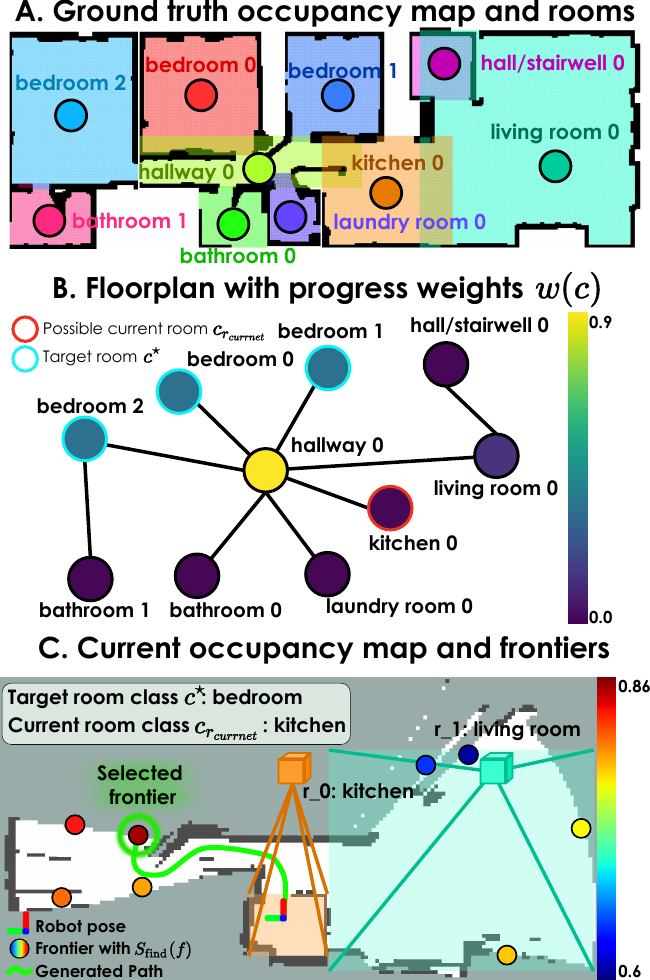}
\caption{\texttt{find\_room} example. The agent is in a kitchen and searches for a bedroom. Since the agent is not localized within the floorplan, all kitchen instances are treated as possible current locations. Neighboring room types are weighted by their shortest graph distance to the target, and these weights are combined with semantic frontier scores to select the exploration goal.}
\label{fig:find_room}
\vspace{-0.5em}
\end{figure}

\mypar{\texttt{find\_room}}
The previous exploration modes operate within parts of the environment that have already been observed. In contrast, this mode is selected when the high-level planner determines that answering the question requires visiting a room type that has not yet been reliably discovered. Rather than searching all unexplored regions uniformly, our method leverages the estimated floorplan graph as a weak topological prior to identify room types that are most likely to lead toward the desired destination. Since the floorplan may be incomplete or contain incorrect room labels or connectivity, it is not treated as an exact map but rather as a source of high-level structure. An overview of the proposed strategy is shown in Fig.~\ref{fig:find_room}. 
The high-level planner first predicts a target room class $\targetclass$ from the floorplan. The label $\class_{r_{current}}$ of the agent's current room node $r_{current}$ is matched against all floorplan nodes with the same semantic class, each of which is treated as a possible current location. The neighboring nodes of these candidate locations define the set of room classes that may be encountered next.

Let $\cU_t \subseteq \nodes{}{\floorplan}$ denote the floorplan nodes adjacent to the current localization candidates, representing rooms that could plausibly be encountered next. We define the corresponding set of candidate room classes as
\begin{equation}
\cC_t = \{\classlabel{u} \mid u \in \cU_t\},
\end{equation}
where $\classlabel{u}$ denotes the semantic room class of floorplan node $u$. For each candidate class $\class \in \cC_t$, we compute its minimum shortest-path distance to the target class $\targetclass$ as
\begin{equation}
\shortestdist{\class}{\targetclass} =
\min_{\substack{
u \in \cU_t:\,\classlabel{u}=\class\\
v \in \nodes{}{\floorplan}:\,\classlabel{v}=\targetclass
}}
\distscore{\graph^{\floorplan}}{u}{v},
\end{equation}
where $\distscore{\graph^{\floorplan}}{u}{v}$ is the shortest-path distance between floorplan nodes $u$ and $v$. Intuitively, $\shortestdist{\class}{\targetclass}$ measures how many room transitions separate a candidate room class from the desired target room type. These distances are converted into normalized progress weights,
\begin{equation}
\weightclass{\class} =
\frac{\exp(-\shortestdist{\class}{\targetclass})}
{\sum_{\class' \in \cC_t}
\exp(-\shortestdist{\class'}{\targetclass})}.
\end{equation}
The resulting weight $\weightclass{\class}$ measures how promising room class $\class$ is as an intermediate step toward the target room type $\targetclass$. Classes closer to the target in $\graph^{\floorplan}$ therefore receive higher weights. \cref{fig:find_room} B. shows an example of floorplan progress weights. If the current room cannot be matched to the floorplan or no path to $c^\star$ exists, $\weightclass{\class}$ are set to zero and the floorplan does not contribute to the exploration gain.

The planner then evaluates each frontier $f_i$ according to
\begin{equation}
\frontiergain{\mathrm{find}}{f_i} = \alpha \frontiergain{\mathrm{tr}}{f_i} + \beta \frontiergain{\mathrm{fp}}{f_i} + \gamma \frontiergain{\mathrm{free}}{f_i}.
\label{eq:find_room_score}
\end{equation}

The transition score $\frontiergain{\mathrm{tr}}{f_i}$ estimates how likely the frontier is to lead to the target room type $\targetclass$. To compute this score, we compare the open-vocabulary frontier feature $\phi_{f_i}$ against the following set of text prompts,
\begin{align*}
\cT(\targetclass)=[\texttt{"doorway to a }\targetclass\texttt{"}, \texttt{"entrance to a }\targetclass\texttt{"}, \\ \texttt{"hallway to a }\targetclass\texttt{"}, \texttt{"opening into a }\targetclass\texttt{"}].
\end{align*}
Let $h$ denote the text embedding of template $\tau_h \in \cT(\targetclass)$. The transition score is defined as the maximum cosine similarity between the frontier feature and the template embeddings,
\begin{equation}
\frontiergain{\mathrm{tr}}{f_i} = \max_{\tau_h \in \cT(\targetclass)} \frac{\phi_{f_i}^\top h} {\norm{\phi_{f_i}}\norm{h}}.
\end{equation}

Intuitively, frontiers whose visual appearance resembles an entrance leading into the target room receive higher scores.

The floorplan-guided term is defined as
\begin{equation}
\frontiergain{\mathrm{fp}}{f_i} = \sum_{c \in \cC_t} \weightclass{\class} \frac{\phi_{f_i}^\top \eta_c} {\norm{\phi_{f_i}}\norm{\eta_c}},
\end{equation}

where $\eta_{\class}$ denotes the text embedding of room class $\class$. Finally, $\frontiergain{\mathrm{free}}{f_i}$ is obtained by normalizing the stored frontier size $s_{f_i}$ across the candidate frontiers.

The highest-scoring frontier is selected as the next exploration target. In practice, this encourages the agent to explore frontiers that appear to lead into the desired room type, while also prioritizing directions that are consistent with the floorplan graph and contain large unexplored regions.

Overall, \method~combines high-level semantic reasoning with specific low-level planning. The \vlm-based task planner determines what information is required to answer the question and selects the exploration mode, while the low-level planner uses the scene graph, semantic frontiers and the floorplan to gather the required information.

%% file: 5-experimental_results.tex

\section{EXPERIMENTS}\label{sec:results}

\subsection{Implementation Details}
Object detection with YOLOe~\cite{wang2025yoloe} uses the HM3D~\cite{Yadav2022HabitatMatterport3S} semantic classes in simulation and the ADE20K~\cite{zhou2019semantic} classes in real-world experiments. We retain at most $M=3$ room key views based on their CLIP similarity to the question. For \textbf{\texttt{explore\_room}}, we set $N=6$; for \textbf{\texttt{go\_to\_objects}}, we set $\epsilon=1$ and $(\visweight{\mathrm{vis}}, \visweight{\mathrm{img}}) = (0.7, 0.3)$ in~\cref{eq:visibility_score}; for \textbf{\texttt{find\_room}}, we set $(\alpha,\beta,\gamma)=(0.35,0.55,0.1)$ in~\cref{eq:find_room_score}. We evaluate GPT-4o, GPT-5.5, Gemini-3.5 Pro, Claude Opus 4.8 and Qwen 3.5-9B as high-level planners in simulation, and use GPT-5.5 for the real-world experiments. We use $\kappa_{\mathrm{min}}=0.8$ in all experiments.

\vspace{-.2em}
\subsection{Simulation Results}

\mypar{Datasets}
Following~\cite{saxena2025GraphEQA,allen2024exploreeqa}, we evaluate \method~using the Habitat~\cite{Savva2019HabitatSim} simulator on HM3D~\cite{Yadav2022HabitatMatterport3S} scenes and the OpenEQA~\cite{OpenEQA2023} and ExploreEQA~\cite{allen2024exploreeqa} question sets. We keep only single-floor episodes with semantic annotations to be able to generate the floorplan priors. This gives 108 OpenEQA episodes and 114 ExploreEQA episodes. For both datasets, we evaluate multiple-choice ($\cA$ is available) and no-choice settings when supported by the method.

\mypar{Floorplan generation}
For each scene, we generate an estimated topological floorplan graph before evaluation. We project HM3D semantic regions onto a bird's-eye-view map, assign each region a semantic room label, and extract room adjacencies from the resulting room map. The resulting floorplan contains only room labels and room connectivity and it does not provide metric geometry or the agent's localization within it. As discussed in~\cref{sec:problem_statement}, the floorplan is treated as a weak structural prior and may contain inaccuracies, including incorrect room labels and connectivity edges.

\begin{table}[!t]
\centering
\setlength{\tabcolsep}{3pt}
\begin{tabular}{lll|ccc|ccc}
\toprule
& \textbf{VLM} & \textbf{Method}
& \multicolumn{3}{c|}{\textbf{OpenEQA}}
& \multicolumn{3}{c}{\textbf{ExploreEQA}}\\
& & & SR\%$\uparrow$ & S$\downarrow$ & P$\downarrow$
& SR\%$\uparrow$ & S$\downarrow$ & P$\downarrow$\\
\midrule

\multirow{15}{*}{\rotatebox{90}{Choices}}
& \multirow{3}{*}{GPT-4o}
& \textbf{HFLEX-EQA} & 52.8 & 12.32 & 31.9 & 48.2 & 15.12 & 40.8\\
& & GraphEQA & 36.1 & 5.44 & 22.3 & 39.5 & 8.12 & 17.5\\
& & ExploreEQA & 44.4 & 13.52 & 24.9 & 44.7 & 11.81 & 22.9\\

\cmidrule(lr){2-9}

& \multirow{3}{*}{GPT-5.5}
& \textbf{HFLEX-EQA} & \best{75.0} & 4.00 & 9.3 & \second{57.9} & 4.47 & 14.4\\
& & GraphEQA & 35.2 & 1.59 & 6.2 & 50.9 & 3.14 & 6.5\\
& & ExploreEQA & 59.3 & 19.11 & 35.0 & 54.4 & 18.74 & 37.5\\

\cmidrule(lr){2-9}

& \multirow{3}{*}{Gem-3.5}
& \textbf{HFLEX-EQA} & \best{75.0} & 4.49 & 13.0 & \best{63.2} & 5.96 & 19.6\\
& & GraphEQA & 46.3 & 1.76 & 4.7 & 54.4 & 2.66 & 5.5\\
& & ExploreEQA & 54.6 & 13.25 & 24.0 & 53.5 & 10.98 & 21.5\\

\cmidrule(lr){2-9}

& \multirow{3}{*}{Opus 4.8}
& \textbf{HFLEX-EQA} & \second{63.9} & 6.12 & 25.6 & \second{57.9} & 5.75 & 23.1\\
& & GraphEQA & 41.7 & 5.09 & 16.1 & 53.5 & 10.76 & 16.3\\
& & ExploreEQA & 54.6 & 15.43 & 28.7 & 43.9 & 15.80 & 32.4\\

\cmidrule(lr){2-9}

& Q3.5-9B
& \textbf{HFLEX-EQA} & 52.8 & 3.54 & 12.8 & 51.8 & 3.98 & 13.6\\
& & GraphEQA & 30.6 & 3.89 & 13.6 & 44.7 & 7.53 & 14.9\\
& & ExploreEQA & 36.1 & 16.12 & 30.3 & 43.0 & 11.40 & 22.3 \\
\midrule

\multirow{8}{*}{\rotatebox{90}{\makecell{No choices}}}
& GPT-4o
& \textbf{HFLEX-EQA} & 36.1 & 16.97 & 40.5 & 36.8 & 17.64 & 64.7\\
& & GraphEQA & 24.1 & 5.19 & 24.2 & 31.6 & 8.05 & 21.7\\

\cmidrule(lr){2-9}

& GPT-5.5
& \textbf{HFLEX-EQA} & \second{44.4} & 3.78 & 12.7 & \second{53.5} & 6.39 & 22.8\\
& & GraphEQA & 26.9 & 1.35 & 5.1 & 50.0 & 3.25 & 6.5\\

\cmidrule(lr){2-9}

& Gem-3.5
& \textbf{HFLEX-EQA} & \best{56.5} & 6.85 & 21.1 & \second{53.5} & 8.89 & 26.3\\
& & GraphEQA & 39.8 & 2.00 & 3.4 & 46.5 & 3.31 & 6.6\\

\cmidrule(lr){2-9}

& Opus 4.8
& \textbf{HFLEX-EQA} & 40.7 & 11.95 & 55.1 & \best{57.9} & 11.28 & 39.7\\
& & GraphEQA & 38.0 & 5.28 & 18.8 & 49.1 & 8.32 & 16.9\\

\cmidrule(lr){2-9}

& Q3.5-9B
& \textbf{HFLEX-EQA} & 37.0 & 11.24 & 29.3 & 36.8 & 13.16 & 33.8\\
& & GraphEQA & 23.1 & 4.12 & 17.6 & 32.5 & 8.89 & 29.4\\

\bottomrule
\end{tabular}
\caption{Main comparison grouped by VLM. SR\% denotes answering success rate, S the average number of planning steps, and P the average path length (m). Q stands for Qwen and Gem for Gemini. \colorbox{bestblue}{\textbf{Best}} and \colorbox{secondorange}{\textbf{second-best}} SR\% for the Choices and No Choices settings are highlighted.}
\label{tab:main_vlm_comparison}
\vspace{0.2ex}

\setlength{\tabcolsep}{5pt}
\begin{tabular}{l|ccc|ccc}
\toprule
\textbf{Config}
& \multicolumn{3}{c|}{\textbf{OpenEQA}}
& \multicolumn{3}{c}{\textbf{ExploreEQA}}\\
&  SR\%$\uparrow$ & S$\downarrow$ & P$\downarrow$ &  SR\%$\uparrow$ & S$\downarrow$ & P$\downarrow$\\
\midrule

\textbf{HFLEX-EQA}
& \best{75.0} & 4.49 & 13.0
& \best{63.2} & 5.96 & 19.6\\

HFLEX-EQA w/o FP
&64.8&6.56&13.1
&53.5&7.53&18.2\\

HFLEX-EQA w/o VP
&\second{67.6} & 7.28 &8.3
&\second{58.8}&7.74&14.5\\

HFLEX-EQA w/o SF
&66.7&12.10&8.6
&57.0&8.80&11.1\\

GraphEQA + FP
&43.5&1.75&6.1
&57.0&3.01&6.8\\

\bottomrule
\end{tabular}

\caption{Ablation study with Gemini-3.5. FP: floorplan prior, VP: viewpoint selection, SF: semantic frontiers (using the closest frontier instead). SR\%: success rate, S: average planning steps, P: average path length (m).}
\label{tab:hvlm_ablation}
\vspace{0.2ex}

\setlength{\tabcolsep}{4pt}
\begin{tabular}{ll|ccc|ccc}
\toprule
& \textbf{Method}
& \multicolumn{3}{c|}{\textbf{OpenEQA}}
& \multicolumn{3}{c}{\textbf{ExploreEQA}}\\
& 
& SR\%$\uparrow$ & S$\downarrow$ & P$\downarrow$
& SR\%$\uparrow$ & S$\downarrow$ & P$\downarrow$\\
\midrule

\multirow{4}{*}{\rotatebox{90}{Choices}}
& \textbf{HFLEX-EQA}
& \second{75.0} & 4.49 & 13.0
& \second{63.2} & 5.96 & 19.6 \\

& \textbf{HFLEX-EQA + GT}
& \best{79.6} & 3.16 & 9.5
& \best{64.0} & 5.11 & 16.6 \\

& GraphEQA
& 46.3 & 1.76 & 4.7
& 54.4 & 2.66 & 5.5 \\

& GraphEQA + GT
& 63.9 & 1.88 & 4.1
& 57.9 & 2.75 & 5.8 \\

\midrule

\multirow{4}{*}{\rotatebox{90}{No choices}}
& \textbf{HFLEX-EQA}
& \second{56.5} & 6.85 & 21.1
& \second{53.5} & 8.89 & 26.3 \\

& \textbf{HFLEX-EQA + GT}
& \best{59.3} & 5.47 & 15.1
& \best{55.3} & 6.82 & 24.2 \\

& GraphEQA
& 39.8 & 2.00 & 3.4
& 46.5 & 3.31 & 6.6 \\

& GraphEQA + GT
& 44.4 & 2.69 & 6.1
& 50.0 & 3.40 & 5.4 \\

\bottomrule
\end{tabular}

\caption{Ground-truth semantics (GT) effect analysis with Gemini-3.5. SR\%: success rate, S: average planning steps, P: average path length (m).}
\label{tab:oracle_semantics}
\vspace{-.5em}
\end{table}

\begin{figure*}[t]
\centering
\includegraphics[width=\linewidth]{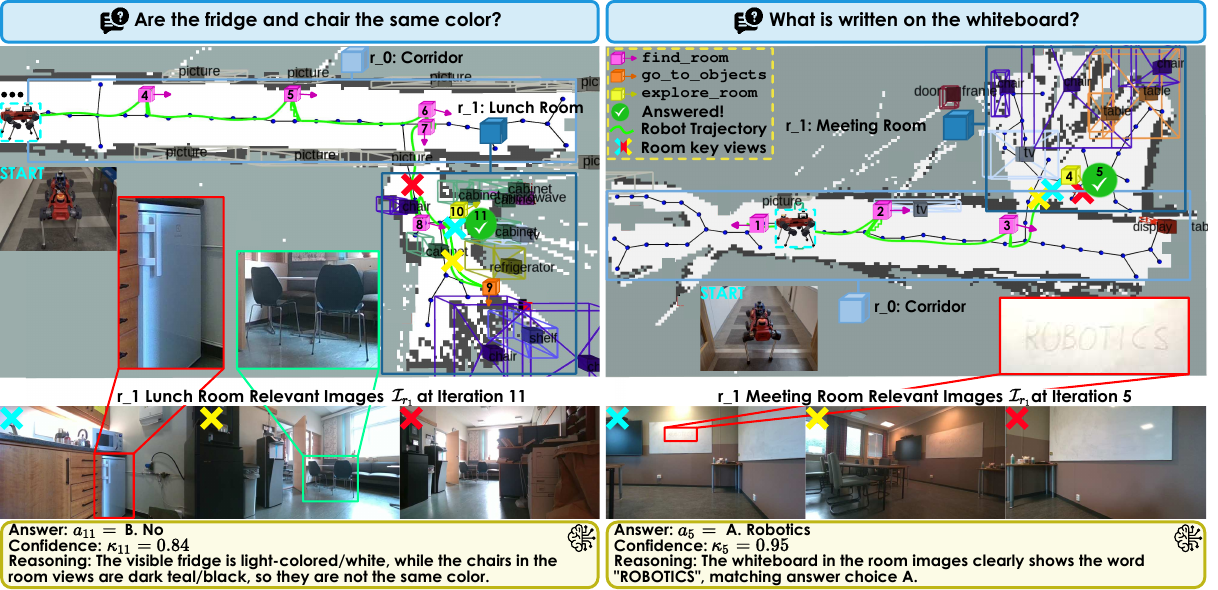}
\caption{Real-world deployment of the proposed framework on a quadruped robot in two \eqa~episodes. The top figures show the constructed occupancy maps and scene graphs, together with the selected exploration mode and goal at each high-level planning iteration. The relevant room images are used to answer the question when the system has gathered enough information.}
\label{fig:experiments}
\vspace{-0.5em}
\end{figure*}
\begin{table*}[t]
\centering
\scriptsize

\begin{tblr}{
  width=\textwidth,
  colspec={
    X[2.0,l]
    X[1.2,l]
    Q[l,wd=22mm]
    |
    X[1.65,c]
    Q[c,wd=7mm]
    Q[c,wd=6mm]
    Q[c,wd=8mm]
    |
    X[2.9,c]
    Q[c,wd=7mm]
    Q[c,wd=14mm]
  },
  cells={valign=m},
  row{1}={font=\bfseries},
  column{1-3}={halign=l},
  column{4-10}={halign=c},
  rowsep=0.8pt,
  colsep=1pt,
  stretch=0.82,
}

\toprule
\SetCell[r=2]{l}\textbf{Question} & \SetCell[r=2]{l}\textbf{Type} & \SetCell[r=2]{l}\vtext{\textbf{Answer}\\\textbf{choices}} & \SetCell[c=4]{c}\textbf{Choices} &  &  &  & \SetCell[c=3]{c}\textbf{No choices} &  &  \\

&  &  & Answer & $\kappa$$\uparrow$ & S$\downarrow$ & P$\downarrow$ & Answer & $\kappa$$\uparrow$ & Equal $m_t$ \\
\midrule

\vtext{What is the\\microwave's color?} & \vtext{State} & \vtext{A. White\\B. Black\\C. Blue\\D. Yellow} & \SetCell{bg=correctgreen,font=\bfseries} A. White & 0.94 & 10 & 48.6 & \SetCell{bg=correctgreen,font=\bfseries} White & 0.93 & 9/10 \\
\midrule

\vtext{How many chairs are\\there to have lunch?} & \vtext{Counting} & \vtext{A. Three\\B. Five\\C. Seven\\D. Two} & \SetCell{bg=correctgreen,font=\bfseries} D. Two & 0.86 & 9 & 42.6 & \SetCell{bg=correctgreen,font=\bfseries} 2 & 0.86 & 9/9 \\
\midrule

\vtext{Are the fridge and chair\\the same color?} & \vtext{Sequential\\understanding} & \vtext{A. Yes\\B. No} & \SetCell{bg=correctgreen,font=\bfseries} B. No & 0.84 & 11 & 50.9 & \SetCell{bg=correctgreen,font=\bfseries} No & 0.88 & 11/11 \\
\midrule

\vtext{What is written on\\the whiteboard?} & \vtext{Identification} & \vtext{A. Robotics\\B. Deep learning\\C. Hello world\\D. University} & \SetCell{bg=correctgreen,font=\bfseries} A. Robotics & 0.95 & 5 & 17.3 & \SetCell{bg=correctgreen,font=\bfseries} ROBOTICS & 0.88 & 5/5 \\
\midrule

\vtext{Is there any visual support\\in the meeting room?} & \vtext{Existence} & \vtext{A. No\\B. Yes} & \SetCell{bg=correctgreen,font=\bfseries} B. Yes & 0.92 & 4 & 16.2 & \SetCell{bg=correctgreen,font=\bfseries} \vtext{Yes, a TV/display and\\a whiteboard} & 0.92 & 4/4 \\
\midrule

\vtext{Where can I\\find my laptop?} & \vtext{Location} & \vtext{A. In the cabinet\\B. On the chair\\C. On the desk} & \SetCell{bg=correctgreen,font=\bfseries} C. On the desk & 0.85 & 4 & 14.6 & \SetCell{bg=correctgreen,font=\bfseries} \vtext{A laptop is on a desk\\in the office off the corridor} & 0.81 & 4/4 \\
\midrule

\vtext{Where can I\\put my jacket?} & \vtext{Functional\\understanding} & \vtext{A. On the hanger in\\the office\\B. In the cabinet\\C. On the chair} & \SetCell{bg=correctgreen,font=\bfseries} \vtext{A. On the hanger\\in the office} & 0.84 & 7 & 42.6 & \SetCell{bg=correctgreen,font=\bfseries} \vtext{On the coat rack/hanger\\in the office} & 0.86 & 5/7 \\
\bottomrule

\end{tblr}

\caption{Real-world \method~evaluation on representative question types. All experiments were executed using the multiple-choice setting. The no-choice results were obtained by replaying the recorded observations, scene graph and planner history while removing the answer choices. $\kappa$ denotes the confidence score of the \vlm~answer, S the number of high-level planning iterations, and P the total path length (m). ``Equal $m_t$'' reports the number of planning iterations in which the selected high-level exploration mode matched the mode selected in the corresponding multiple-choice run.}
\label{tab:real_experiments}
\vspace{-3em}
\end{table*}

\mypar{Baselines}
We compare \method~against the \vlm-based ExploreEQA~\cite{allen2024exploreeqa} and GraphEQA~\cite{saxena2025GraphEQA}. ExploreEQA scores images and proposes image-space frontiers that are projected into a semantic occupancy map, while GraphEQA provides a scene graph and question-relevant images to the \vlm. ExploreEQA is evaluated only with answer choices, as its answer is based on comparing probabilities over a fixed answer set across all planning iterations and does not support free-form answers. All methods use a budget of 50 iterations.

\mypar{Metrics}
We report success rate (SR\%), the average number of planning steps (S), and path length in meters (P).
For multiple-choice runs, success means selecting the correct answer. For no-choice runs, answers are graded by an \llm. 

\mypar{Comparison with baselines}
A quantitative comparison against the proposed baselines is reported in~\cref{tab:main_vlm_comparison}. Across all evaluated \vlm s, \method~consistently achieves the highest success rate on both datasets. The largest improvements are obtained with the stronger language models, where \method~reaches up to $75.0\%$ success rate on OpenEQA and $63.2\%$ on ExploreEQA. These improvements demonstrate that combining a scene graph representation with hierarchical planning, leveraging a structural floorplan graph, enables us to achieve \sota results on the \eqa~task. In contrast, ExploreEQA generally requires substantially more planning iterations and longer trajectories, while GraphEQA typically terminates after fewer planning steps but at the cost of lower success rates. Although GraphEQA often needs fewer planning iterations, this is largely because it finishes exploration earlier. The higher success rates achieved by \method~indicate that the additional planning iterations are spent gathering more question-relevant information.

\mypar{Ablation study}
The ablation study in~\cref{tab:hvlm_ablation} confirms that each proposed component contributes to the final method performance. Removing the floorplan prior consistently reduces success rates, demonstrating the benefit of exploiting a weak global structural prior during exploration. Similarly, replacing the viewpoint selection or semantic frontiers degrades performance. Finally, adding the floorplan graph to GraphEQA's \vlm~prompt provides mixed results, decreasing performance on OpenEQA while slightly improving ExploreEQA. This suggests that simply providing the floorplan to the prompt does not consistently improve performance.

\mypar{Ground truth semantics analysis}
Results using ground truth semantic information are shown in~\cref{tab:oracle_semantics}. Both \method~and GraphEQA benefit from ground-truth semantics. Nevertheless, \method~outperforms GraphEQA, showing that the gains persist when semantic perception errors are reduced.

\subsection{Real-world Results}

We deploy our framework on a quadruped robot equipped with a custom sensing and computing payload. The platform uses an Ouster OS0 LiDAR and a VectorNav VN100 \imu~for odometry, an Intel RealSense D455 RGB-D camera, and an NVIDIA Jetson Thor for onboard computation. All perception, mapping, and hierarchical planning are executed onboard in real time, except for the \vlm s, which are queried via their APIs.

Table~\ref{tab:real_experiments} and Fig.~\ref{fig:experiments} demonstrate the deployment of \method~in previously unseen indoor environments. Our method successfully answers questions about object state understanding, counting, localization, functional understanding, text recognition, and sequential reasoning. The examples in Fig.~\ref{fig:experiments} illustrate how the high-level planner switches between different exploration modes to gather the information required to answer each question. To evaluate the effect of providing answer choices to the \vlm, we replay the recorded observations and scene graph, and feed them to our method without providing the answer choices. As shown in~\cref{tab:real_experiments}, the free-form answers remain correct and 47/50 planning modes match the multiple-choice runs, showing that planner decisions are largely consistent without answer choices.

Additionally, we test the \vlm~planner's robustness by replaying the real-world episodes with perturbed images while keeping the scene graph and planner history unchanged. We apply Gaussian blur, illumination degradation, and gamma correction with increasing severity. The results are shown in~\cref{fig:real_robustness}. Mode consistency measures how often the planner selects the same exploration mode as in the original run, while answer success measures if the final answer remains correct. Mode consistency remains stable across perturbations, while answer success decreases under strong blur.

%% file: 6-conclusions.tex
\section{CONCLUSIONS}\label{sec:conclusions}

We propose \method, a hierarchical framework for \eqa~in previously unseen indoor environments. Our approach combines an online hierarchical scene graph, \vlm-based planning and semantic frontier exploration enhanced with a topological floorplan graph. By decomposing exploration into room exploration, object inspection, and room finding, \method~selects exploration strategies based on the information required to answer each question and the current scene graph. Extensive experiments on the OpenEQA and ExploreEQA benchmarks show \method~outperforms existing \eqa~methods, while real-world experiments demonstrate successful deployment on a quadruped robot. These results demonstrate the benefits of combining hierarchical planning, structured scene representations, and floorplan priors for the \eqa~task.

\begin{figure}[t]
\centering
\includegraphics[width=\linewidth]{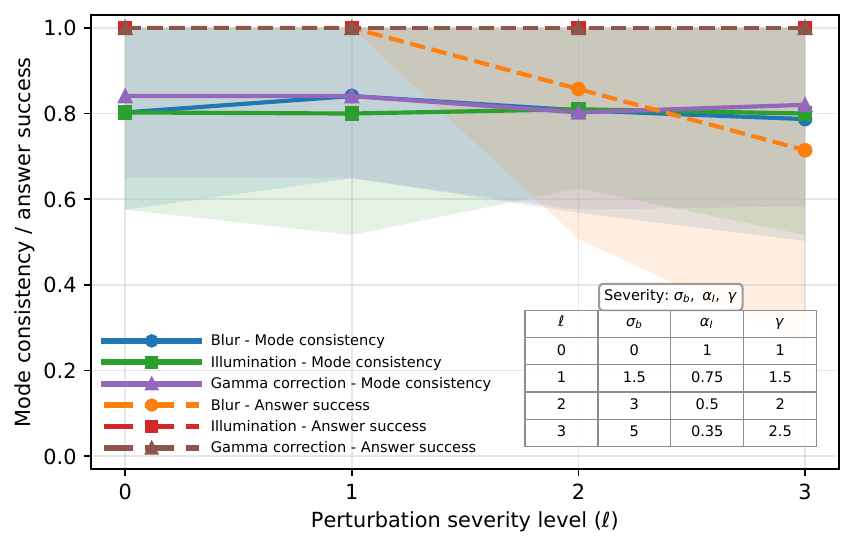}
\vspace{-2em}

\caption{Robustness analysis on real-world data. Mode ($m_t$) consistency and answer success under visual perturbation are shown. The table maps each level $\ell$ to the blur strength $\sigma_b$, illumination scale $\alpha_I$, and gamma value $\gamma$.}
\label{fig:real_robustness}
\vspace{-0.5em}
\end{figure}